%% file: ms.tex
\documentclass{article} 
\usepackage{iclr2022_conference,times}

\input{math_commands.tex}

\usepackage{url}
\usepackage{hyperref}
\usepackage{graphicx}
\usepackage{cleveref}
\usepackage{todonotes}
\setuptodonotes{inline}
\usepackage{tikz}
\usepackage{caption}
\usepackage{subcaption}
\usepackage{gensymb}

\definecolor{blue1}{RGB}{0,102,189}
\definecolor{blue2}{RGB}{98,160,214}
\definecolor{my_orange}{RGB}{243,98,33}
\definecolor{mygreen}{RGB}{30,106,57}
\definecolor{mypurple}{RGB}{91,39,125}
\definecolor{myred}{RGB}{241,13,12}

\input{tikzstyles.tex}

\title{A Deep Learning Approach for Thermal Plume Prediction of Groundwater Heat Pumps}

\author{Raphael Leiteritz, Kyle Davis, Miriam Schulte, \& Dirk Pfl\"uger  \\
Institute for Parallel and Distributed Systems \\
University of Stuttgart\\
Stuttgart, Germany \\
\texttt{\{raphael.leiteritz,kyle.davis\}@ipvs.uni-stuttgart.de} \\
}

\iclrfinalcopy 
\begin{document}

\maketitle

\begin{abstract}
   Climate control of buildings makes up a significant portion of global energy consumption, with groundwater heat pumps providing a suitable alternative.
   To prevent possibly negative interactions between heat pumps throughout a city, city planners have to optimize their layouts in the future.
   We develop a novel data-driven approach for building small-scale surrogates for modelling the thermal plumes generated by groundwater heat pumps in the surrounding subsurface water.
   Building on a data set generated from 2D numerical simulations, we train a convolutional neural network for predicting steady-state subsurface temperature fields from a given subsurface velocity field.
   We show that compared to existing models ours can capture more complex dynamics while still being quick to compute.
   The resulting surrogate is thus well-suited for interactive design tools by city planners.
\end{abstract}

\section{Introduction}
\label{sec:intro}

Heating and cooling of buildings has garnered attention in recent years, especially as the world moves towards renewable sources of energy. 
One focus recently has been on using shallow geothermal energy through groundwater heat pumps (GWHP) \citep{Halilovic2022}. 
As the groundwater temperature is relatively stable year round, GWHPs are able to use this source to both heat and cool buildings. 
As cities move towards installing more GWHPs, assessing and optimizing their influence on the subsurface is required. 
Installing GWHPs without any restrictions could result in situations where negative interaction occurs \citep{Garcia2020, Daemi2019}. 
GWHPs operate by extracting water from an extraction well, either heating or cooling this fluid, and re-injecting it back into the subsurface.  
The local temperature changes around the injection well, and a thermal plume develops due to the diffusion and advection of this water in the subsurface. 
This thermal plume can propagate downstream and interact with other GWHPs or even recirculate into its own extraction well, causing interference. 
This requires careful planning of the layout and operational loads of GWHPs \citep{Beck2013}. 
To provide an accurate assessment of the groundwater temperatures due to the usage of many heat pumps, high-fidelity subsurface flow simulations are required \citep{Meng2019}. 
Furthermore, to optimize the layout and usage of a large number of GWHPs on this scale requires many high-fidelity simulation runs, making large optimization scenarios infeasible.

A common solution is to use surrogate models, also known as low-fidelity models, which are computationally cheap to solve the optimization problem \citep{Sbai2019, Nagoor2019, Robinson2012}. 
To help optimize the layout of potentially thousands of GWHPs, surrogate models could be used to determine the local temperature influence that each GWHP has on the groundwater temperature. 
As the thermal influence of one heat pump can be computed fast and cheaply, many such evaluations can be performed to determine the influence of multiple heat pumps in a system. 
This can be used to provide either a semi-optimised solution, or to evaluate the influence of one heat pump on other previously installed heat pumps. 

\begin{figure}[!htb]
\centering
\begin{subfigure}{.5\textwidth}
  \centering
  \includegraphics[width=.8\linewidth]{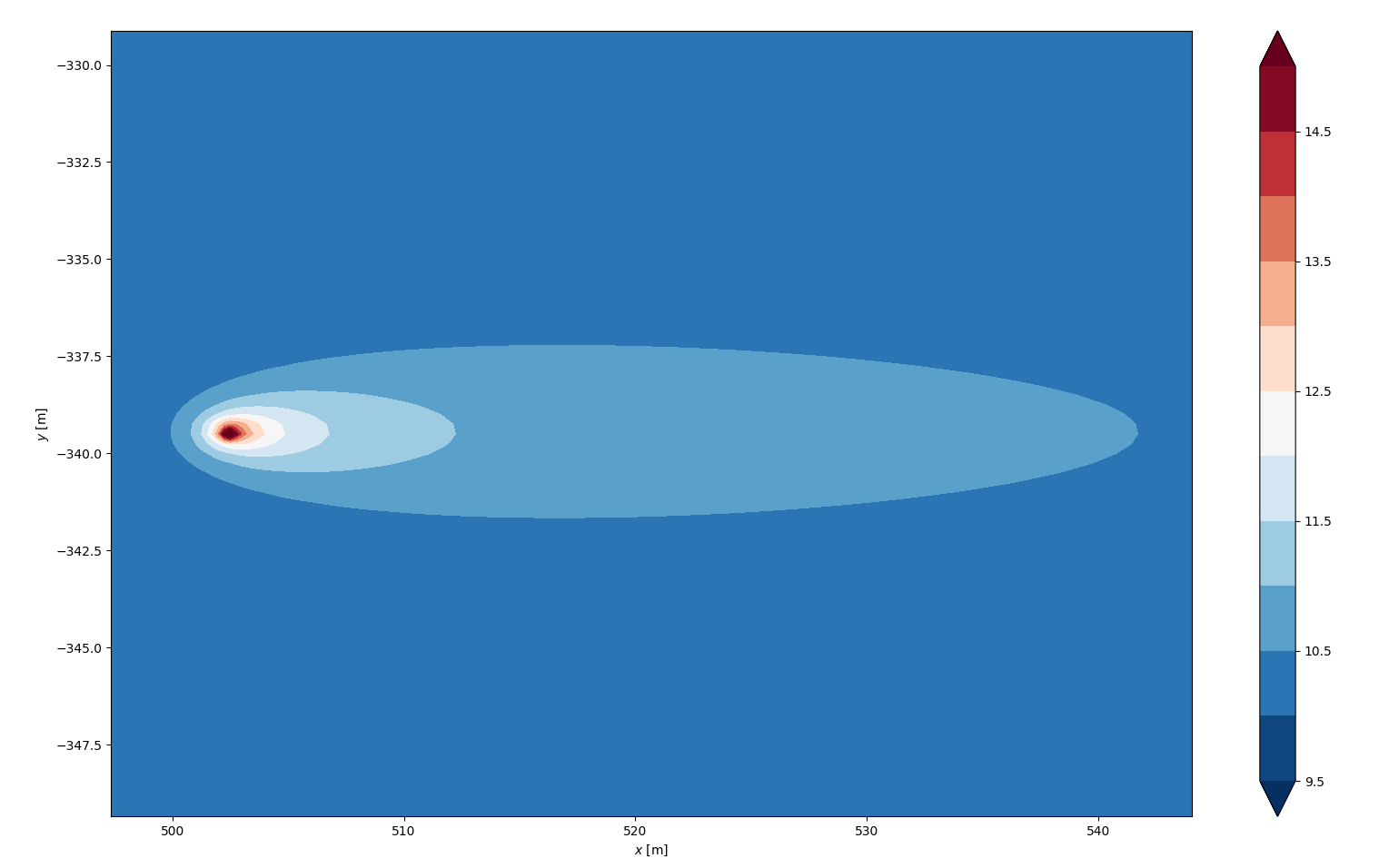}
  \caption{Analytical thermal plume}
  \label{analyticalPlume}
\end{subfigure}%
\begin{subfigure}{.5\textwidth}
  \centering
  \includegraphics[width=.8\linewidth]{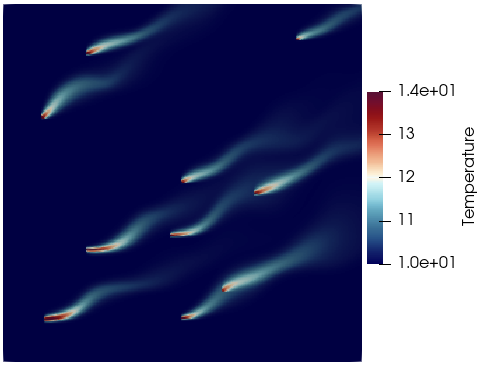}
  \caption{Temperature field with 10 GWHPs scattered throughout the domain.}
  \label{large_example}
\end{subfigure}
\caption{The thermal plume developed by (a) GWHP derived from the LAHM model \citep{Pophillat2020} and (b) 10 GWHPs in a large domain using a spatially varying permeability field. The thermal plume in (a) is uni-directional for the analytical model, and is dependent only on the parameters at the location of re-injection. A more realistic example in (b) shows that the thermal plumes follow the velocity direction of the subsurface.}
\label{fig:plume_and_overview}
\end{figure}

Analytical solutions provide a computationally cheap solution to predict the thermal plume \citep{Pophillat2020}, but suffer from several disadvantages. 
They often do not account for variations in the groundwater parameters in space, such as varying permeability field, pressure gradients and velocity fields. 
This results in the thermal plume extending in one direction only for the analytical solutions (Fig.~\ref{analyticalPlume}), however, the thermal plumes may actually change their direction due to heterogeneous groundwater parameters or obstructions. 

In a typical application the optimization problem would consist of an area with many GWHP's (such as in Fig.~\ref{large_example}, and potentially much larger areas), with known geological parameters; spatially varying permeability field and pressure boundary conditions. 
The surrogate model must be able to provide the local temperature variation (thermal plume) around each individual GWHP (Fig.~\ref{fig:sub1_temp}), and we can determine which GWHPs might interact with each other.

\subsection*{Related Work}
\todo[disable]{extend, less PINN focus}
The usage of machine learning based surrogates for physical applications has increased dramatically in recent years \citep{Vinuesa2021,Chen2021,Zhu2019}.
There have been purely supervised learning approaches put forth \citep{nn_pde_data} with multiple variations using e.g.\ convolutional neural networks (CNNs) \citep{nn_pde_cnn, Thuerey2019} or LSTMs \citep{dl_lstm} for temporal modelling.
Another promising line of research are physics-informed neural networks (PINNs), where a neural network is trained to satisfy some specific boundary-value problem \citep{pinn,pinn_subsurface,pinn_hfm,pinn_pathologies,Gao2021}. 
Given a partial differential equation (PDE), the residual is added to the network's loss function which is then trained to minimize this residual.
Other works follow a coupled approach such as CFDNet \citep{cfdnet} where a neural network surrogate is used as a pre-conditioner inside a traditional numerical solver or e.g. \citet{BECK2019108910} where a neural network is used to model closure terms for turbulent flows.
In this work, we focus on a purely supervised CNN approach for making new predictions by focusing on a very confined flow domain.

\section{Method}
\label{sec:method}



The surrogate model only needs to solve for the local temperature field around a GWHP. 
We assume that the subsurface temperature is domniated by the advection term in the Darcy flow equations (Eq.~\ref{eq:energy}), and the temperature profile follows the velocity field profile. 
Therefore, our aim is to develop a surrogate model that uses the spatially varying, steady-state groundwater velocity field as an input, and outputs the thermal plume that develops due to the addition of a single GWHP in a smaller domain.

\subsection*{Data Generation}

In order to train the neural network, we need to provide enough training data such that the thermal plume can be accurately captured. We use PFLOTRAN \citep{pflotran-paper} to perform high-fidelity subsurface simulations.
We solve for the boundary value problem using a finite volume subsurface solver to determine the pressure field, velocity field and temperature field.
We consider a 2D domain with directions $x$ and $y$, and Darcy velocities $\vq = (q_x, q_y)$.
The training data simulations are performed on a 64x64 structured grid with 4096 cells, covering an area of 128$m$ $\times$ 128$m$ $\times$ 1$m$. The GWHP is modelled by injecting fluid of 0.05 $kg \cdot s^{-1}$ at 15 \degree C at the center of the domain, and run for a total of 720 days to achieve a pseudo steady-state solution. The domain temperature is set to 10 \degree C at time T $=$ 0.

\begin{figure}[!htb]
\centering
\begin{subfigure}{.33\textwidth}
  \centering
  \includegraphics[width=.8\linewidth]{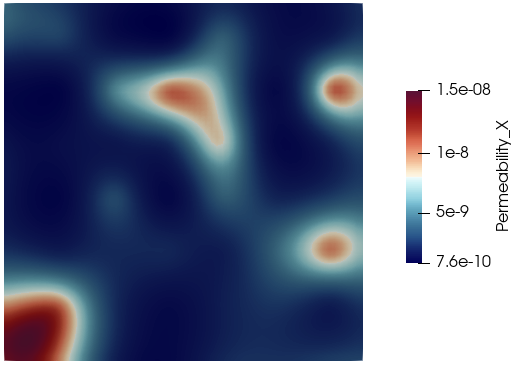}
  \caption{Permeability Field}
  \label{fig:sub1_perm}
\end{subfigure}%
\begin{subfigure}{.33\textwidth}
  \centering
  \includegraphics[width=.8\linewidth]{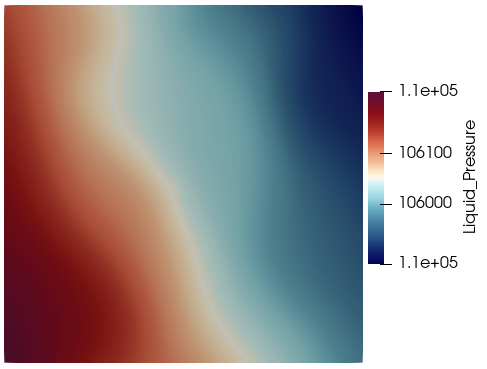}
  \caption{Pressure Field}
  \label{fig:sub1_pressure}
\end{subfigure}
\begin{subfigure}{.33\textwidth}
  \centering
  \includegraphics[width=.8\linewidth]{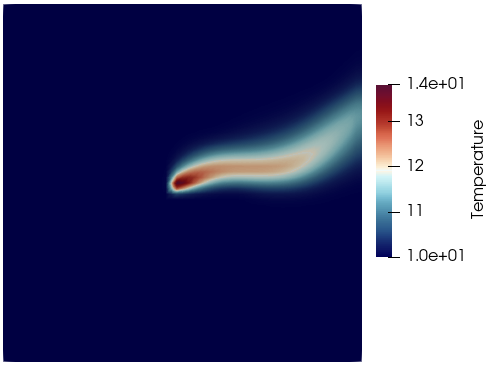}
  \caption{Temperature Field}
  \label{fig:sub1_temp}
\end{subfigure}%
\caption{The permeability field (a), pressure field (b) and temperature field (c) for a training data example. The temperature plume that develops from the GWHP is not uni-directional like the analytical solutions, and depends on the velocity magnitude and direction.}
\label{fig:data_example}
\end{figure}

The training data is generated by varying the permeability field and pressure boundary conditions to generate randomly varying velocity fields.
The permeability field is generated by randomly assigning uniformly distributed values between $4.1 \cdot 10^{-8}$ and $2.1 \cdot 10^{-9}$ at various points on a $4 \times 4$, $6 \times 6$ and $8 \times 8$ square grid throughout the domain. We use the Python library \textit{random} to pseudo-randomly generate the permeability values. These values are then mapped to the PFLOTRAN grid using that radial basis function (RBF) interpolation method with global thin-plate-splines basis functions.
An example permeability field is shown in Fig.~\ref{fig:sub1_perm}.

The pressure gradient applied to the domain is also generated using the \textit{random} library in Python. Two random values for the $x$ and $y$ direction of the gradient are applied.
This allows us to generate many varying velocity fields, and due to the small size of the FV solver, the simulations to calculate a stead-state temperature field $T$ are computationally feasible.
For more details regarding the simulation setup see Appendix \ref{ap:sim_details}.

\todo[disable]{underlying equations are the (stationary) Richards flow equations}
\todo[disable]{generate arbitrary configurations with smooth random flow fields}
\todo[disable]{injection in the middle of the domain leads to some temperature plume}
\todo[disable]{we solve a bunch of these until a quasi-stationary state is achieved}

\subsection*{Preprocessing}
Before using the data we apply some preprocessing to aid the training process.
First, we subtract $10 \degree C$ from all temperature values as this is the initial groundwater temperature we set for our simulation.
We then normalize each quantity (Darcy flow $(q_x, q_y)$ and temperature $T$) separately by centering and scaling the data such that they are restricted to the $[-1, 1]$ range.
This often simplifies the learning task for NNs as has been shown previously e.g.\  in \citep{LeCun2012,Wiesler2011}.
Additionally, we augment and extend the data set by appending randomly rotated input and target images resulting in a larger data set for more robust results.

\subsection*{Model}
\todo[disable]{as surrogate we use a TurbNet structure}
\begin{figure}[!htb]
   \centering
   \includegraphics[width=0.8\textwidth]{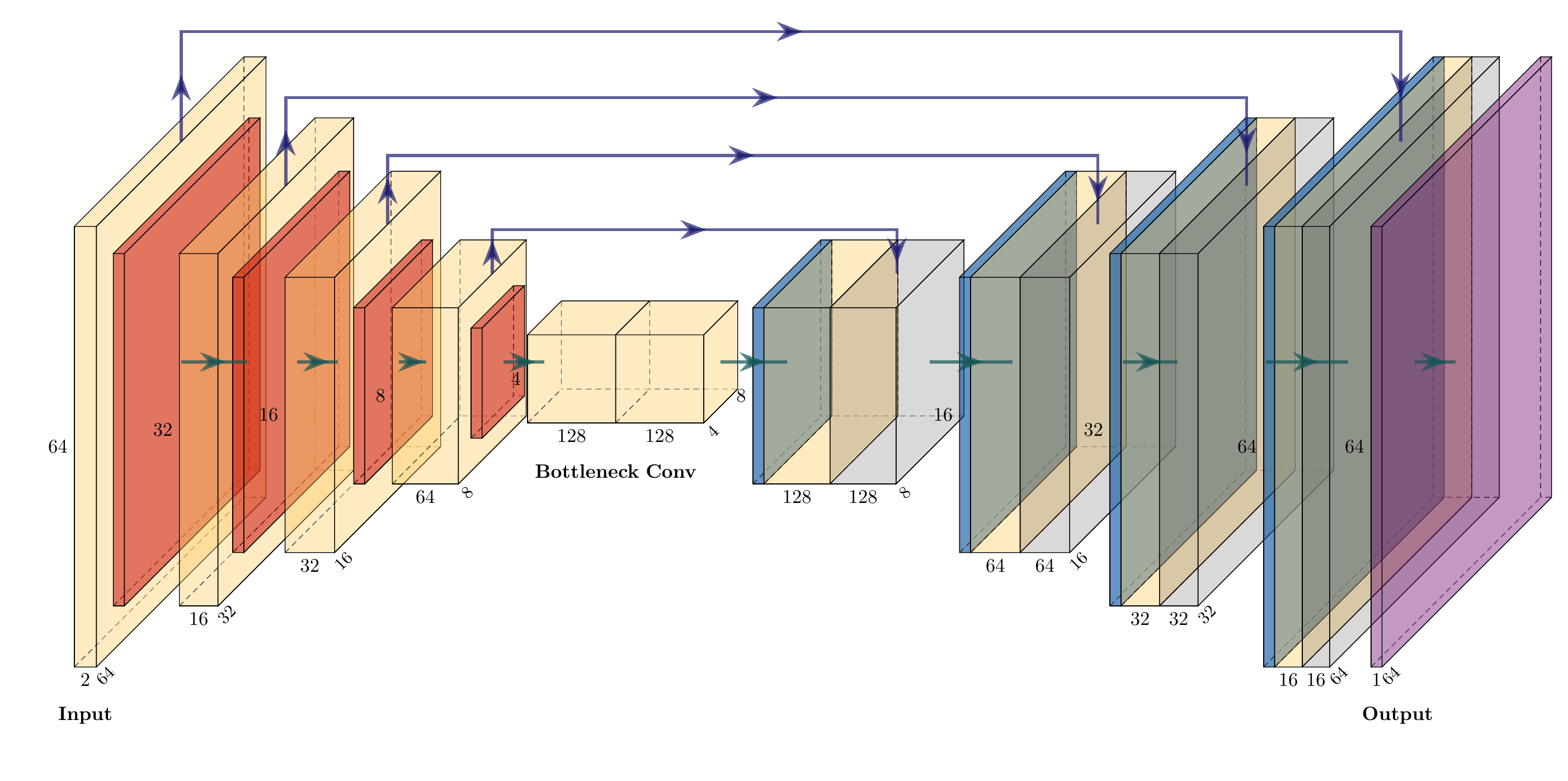}
   \caption{Visualization of the U-net type network architecture. The encoder and decoder paths feature 2D convolutions and ReLU activation functions. Both paths are connected via skip connections in each level.}
   \label{fig:arch}
\end{figure}

For the surrogate model, we choose a slightly modified version of the ``TurbNet'' architecture as described in \citet{Thuerey2019} which in turn can be considered a variant of the ``U-net'' \citep{Ronneberger2015}.
The input is given as a two-channel image with $64\times64$ pixel values corresponding to the Darcy flow $\vq$.
The general shape is similar to a simple ``U-net'' where the input is convolved into coarser and coarser features on a contracting path until a bottleneck level is reached. 
From this on the feature channels are again de-convolved in a symmetric fashion upwards.
Additionally, each step in the hierarchy also includes a skip connection, copying intermediate results from the contracting path and concatenating it feature-wise to the results of the expanding path.
As output, the network produces a single channel image at the same $64\times64$ resolution of the predicted temperature field.
A graphical overview of the architecture is depicted in Fig.~\ref{fig:arch}.
In contrast to \citep{Thuerey2019} we raised the feature size of the bottleneck path.
Our assumption is that features which are convolved down to a single pixel value don't hold much relevant information anymore, and we can thus save some model complexity.
Additionally, we lowered the amount of channels in each block as our experiments showed good results even with fewer parameters.
This approach allows us to directly predict a stationary temperature plume for any given flow field.

\todo[disable]{why did we choose this?}
\todo[disable]{- allows us to directly predict stationary temperature plume for a given flow field}
\todo[disable]{- can then be used in a real-time setting to make predictions for the website user}

\section{Results}
\label{sec:results}


\begin{figure}[!htb]
   \centering
   \begin{subfigure}{0.9\textwidth}
      \includegraphics[width=\textwidth]{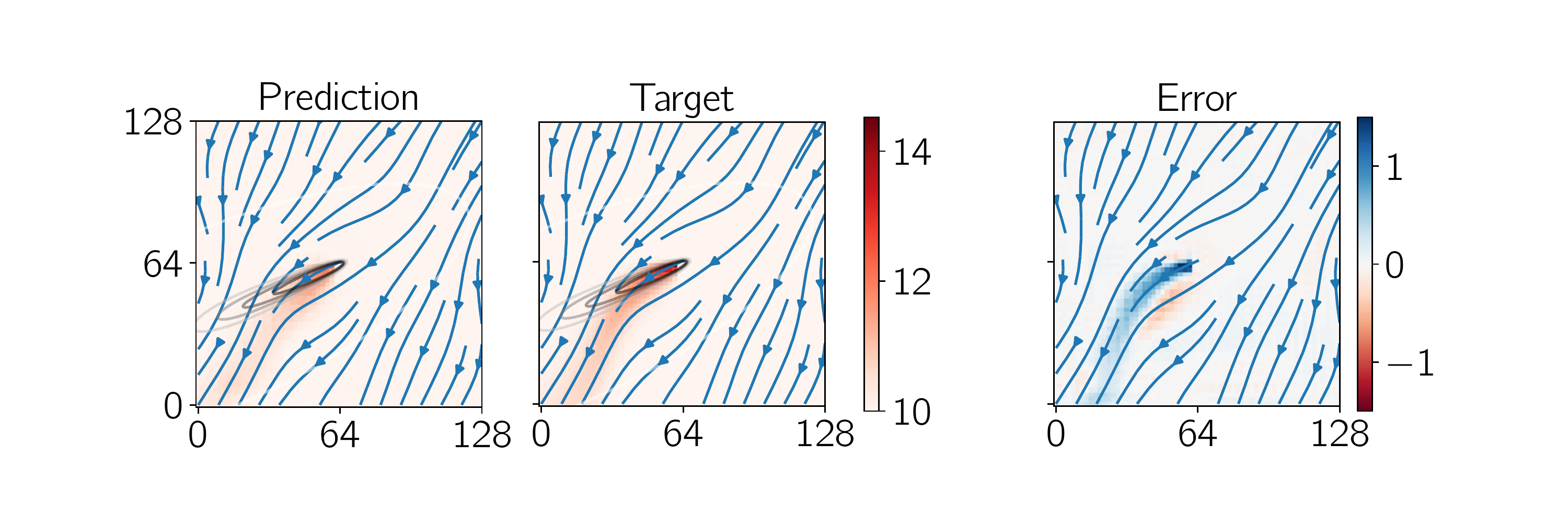}
      \vspace{-1cm}
   \end{subfigure}
   \begin{subfigure}{0.9\textwidth}
      \includegraphics[width=\textwidth]{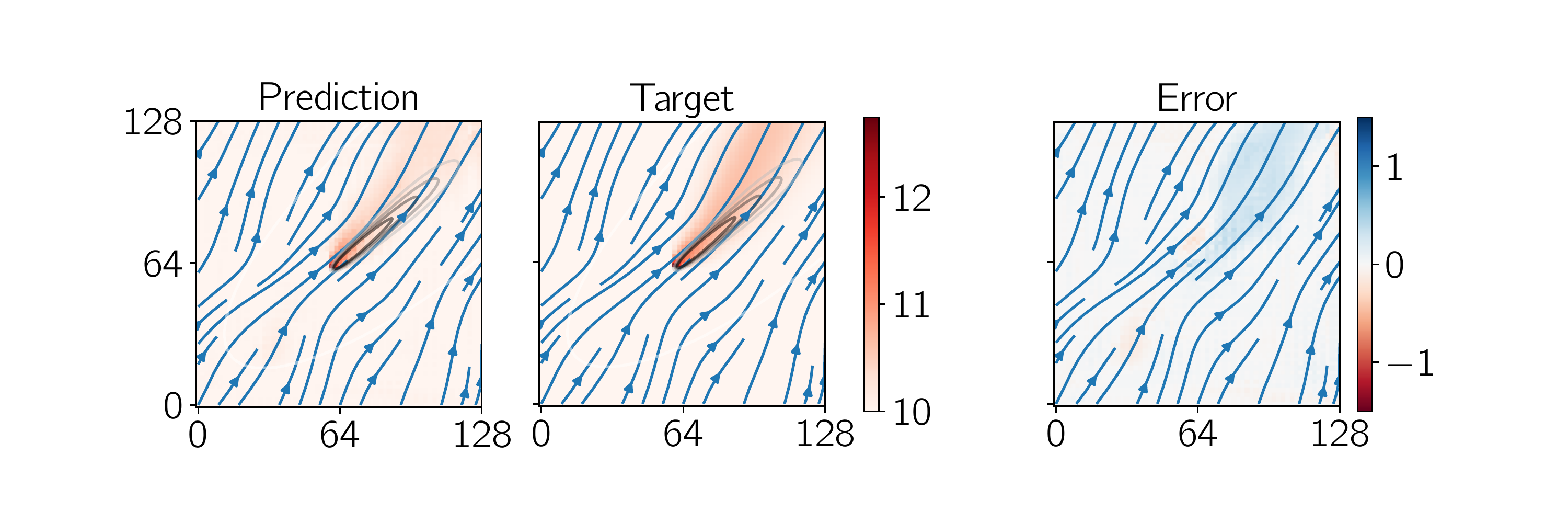}
      \vspace{-1cm}
   \end{subfigure}
   \begin{subfigure}{0.9\textwidth}
      \includegraphics[width=\textwidth]{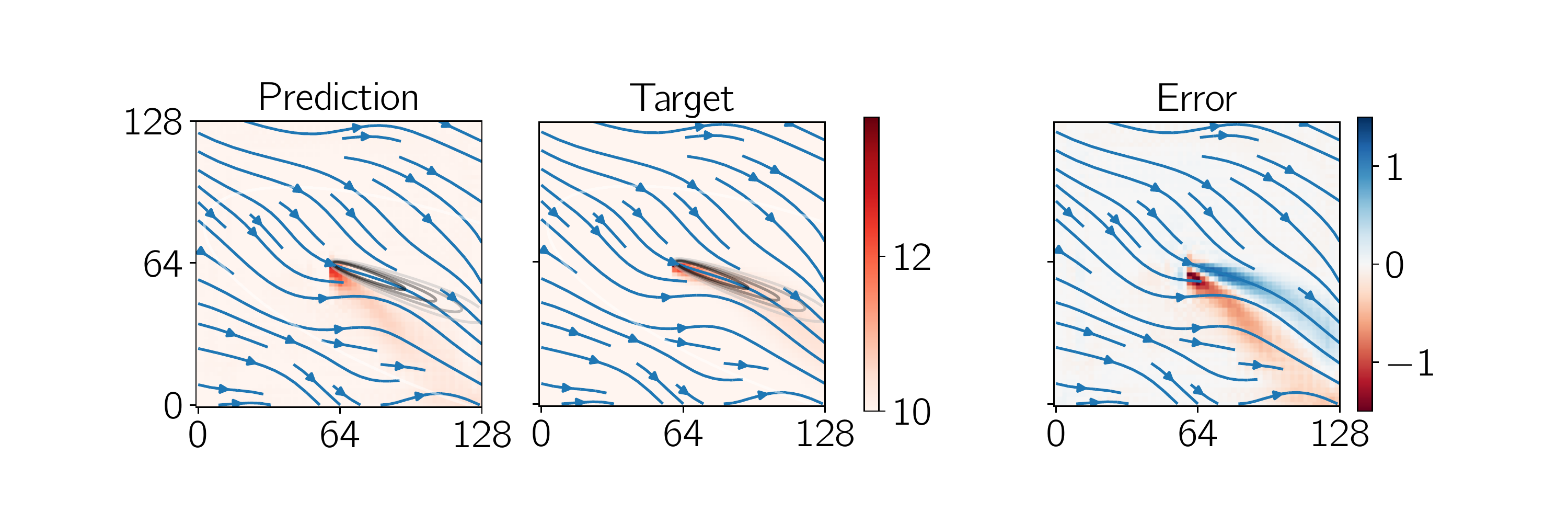}
      \vspace{-1cm}
   \end{subfigure}
   \caption{Prediction result and simulation ground truth for three different samples from the test data set. The predicted temperature plumes match the target in direction and shape. We show two typical results and one outlier. The rightmost column displays the error in \degree C. Note the different scales of the error. Analytical temperature plumes calculated with the LAHM model are overlaid on top in gray.}
   \label{fig:results}
\end{figure}


We trained the aforementioned model on a data set of $239$ generated samples. An additional $720$ augmented samples are created by randomly rotating the input and output images resulting in a total data set size of $959$. $192$ samples are randomly selected as validation data during training.
The model is trained using the ADAM optimizer \citep{adam} with a fixed learning rate $0.0004$ for $60,\!000$ epochs.
For further details see Appendix \ref{ap:training_details}.
Figure~\ref{fig:results} shows a selection of three different predictions of the trained model which were performed using a held out test set of size $40$ which was never before seen by the model or used during training in any way.
Shown in each row are the model's prediction, the target temperature field from the numerical solution and the point-wise error between the two.
Furthermore, the analytical LAHM model from \citet{Pophillat2020} is overlaid on top of both prediction and target plots.
We can nicely observe that in all cases the predicted temperature plume matches the target in direction and shape.
The top example prominently displays the advantage of our method when compared to the analytical model.
While the analytical plume only estimates linear transport of the flow, our method provides a much more detailed prediction following the flow.
The second row again is a nice example of the prediction matching the target quite well also in magnitude.
Lastly, we also include an example where the model did not perform too well and deviates from the expected flow direction showing the predictions are not fully robust yet.
Quantifying the error we see each example having different magnitudes ranging from $0.25$\,\degree C to $2$\,\degree C.
Note however, that the largest error is often only encountered at the injection site with the rest of the plume showing significantly less errors. 
This shows it is harder to determine the initial injection temperature from the velocity magnitude at the GWHP, and deserves more attention in the future. 
The relative error
\begin{equation*}
    \frac{\sum_i \left| T^p_i - T^t_i \right|}{\sum_i \left| T^t_i \right|}
\end{equation*}
between the predicted temperatures $T^p$ and target temperatures $T^t$ achieved on the complete test data set is $0.68 \%$.
\todo[disable]{this is not so bad, rel. error over whole data set 1\% and high errors mostly due to "small" misalignment of plume}
Considering that the analytical model plume can be offset from the real plume by $> 10m$ at the end of the plume, this could make the difference in a real-world scenario when deciding whether it is feasible to place a GWHP on a specific building or not.
\todo[disable]{- evaluation (error on test set, inference time)}
\todo[disable]{e.g. in a real-world scenario a bending plume shifting by ~10m could make the difference when deciding where to place gwhps}
\todo[disable]{- overlay of local surrogates on city map (?)}

\section{Conclusion}
\label{sec:conclusion}

\todo[disable]{conclusion here, demonstrate feasibility of the method, confident that more compute/examples will further improve in the future}
\todo[disable]{mention inference time? with only ~0.1s this allows real-time applications}
\todo[disable]{- future work: overlapping local predictions $\rightarrow$ resolve with global city-wide model}
In this work we have demonstrated the viability of data-driven surrogate models for approximating subsurface thermal plumes generated by groundwater heat pumps.
With a supervised learning approach and a suitable CNN architecture we were able to get a good agreement between fast surrogate predictions and high-fidelity ground truth data.
Especially compared to simpler analytical approaches the data-driven model has shown very convincing results.
This study can be seen as a first step enabling more refined approaches and applications down the line.

Considering the low inference time of the trained model ($\leq 50ms$) we could imagine it being used in close-to-real-time applications such as design tools for city planners.
We would also like to expand the surrogate model by including parametric inputs such as a heat pump's energy output.
Furthermore, a shortcoming of the current approach is the limitation to a small constrained domain with only a single heat pump.
To model larger scenarios with multiple heat pumps another extension could be stitching together multiple local surrogate predictions and then using these as a preconditioned starting point for solving larger numerical models on the scale of a whole city domain.

\todo[disable]{Web app currently uses analytical model. Want to add this as it uses non-uniform velocities.}

\subsubsection*{Acknowledgments}
Funded by Deutsche Forschungsgemeinschaft (DFG, German Research Foundation) under Germany's Excellence Strategy - EXC 2075 – 390740016. We acknowledge the support by the Stuttgart Center for Simulation Science (SimTech).

\bibliography{iclr2022_conference}
\bibliographystyle{iclr2022_conference}

\appendix

\section{Training Details}
\label{ap:training_details}
The model was implemented and trained using \emph{PyTorch} \citep{pytorch} on a workstation with two NVIDIA RTX 3090 GPUs. The batch size was fixed at $64$ samples and the ADAM optimizer was used with a learning rate of $4e$-$4$ and otherwise default parameters.
For supervision a simple MSE loss function is employed.
The model used to produce the results in Sec.~\ref{sec:results} features a total of $480$k parameters.
\todo[disable]{add anonymous github repository}


\section{Simulation Details}
\label{ap:sim_details}
The PFLOTRAN \citep{pflotran-paper} package is used to solve a steady-state thermal-hydrological problem with equations for mass and energy balance given by
\begin{equation}
   \label{eq:mass}
   \nabla \cdot \left(\eta \vq \right) =  \nabla \cdot \left(\eta  K \nabla P  \right) = Q_w
\end{equation}
and
\begin{equation}
   \label{eq:energy}
   \nabla \cdot \left(\eta \vq H\right) - \kappa \Delta T = Q_e
\end{equation}

where $Q_w$ is the mass flow source term, which is non-zero only at the boundaries and at the center where mass is injected into the groundwater by the GWHP, the energy source term $Q_e$, enthalpy $H$, thermal conductivity $\kappa$ and temperature $T$. The relevant parameters are shown in Table \ref{tab:params}.

\begin{table}[ht]
\caption{Simulation parameters}
\centering
\begin{tabular}{| c | c | c |}
   \hline
   $\eta$   & molar density        & $55.3454010547$ $kmol/m^3$ \\
   \hline
   $\kappa$ & thermal conductivity & $0.5 W/(mK)$                   \\
   \hline
   H        & specific enthalpy of water at 10\degree C         & $1.134945 kJ/mol$                  \\
   \hline
\end{tabular}
\label{tab:params}
\end{table}


\todo[disable]{- using these we can additionally regularize the network during training}
\todo[disable]{- makes sure that physically plausible solutions are found}
\todo[disable]{- should be more robust?}
\todo[disable]{- comparison with and without physical term (speeds up training?)}

\end{document}

%% file: math_commands.tex
\usepackage{amsmath,amsfonts,bm}

\def\eqref#1{equation~\ref{#1}}

\def\1{\bm{1}}

\def\vq{{\bm{q}}}

\DeclareMathAlphabet{\mathsfit}{\encodingdefault}{\sfdefault}{m}{sl}
\SetMathAlphabet{\mathsfit}{bold}{\encodingdefault}{\sfdefault}{bx}{n}



%% file: tikzstyles.tex
\usepackage{tikz-dimline}

\tikzset{inNode/.style={draw=incolor, fill=incolor, circle, inner sep=1.5pt}}
\tikzset{outNode/.style={draw=outcolor,  fill=outcolor, circle, inner sep=1pt}}
\tikzset{projNode/.style={draw=outcolor, fill=outcolor, circle, inner sep=1pt}}

\usetikzlibrary{shapes.arrows, fadings, calc, positioning}
\tikzfading[name=fade right,
  left color=transparent!0, right color=transparent!100]
\tikzfading[name=fade left,
  left color=transparent!100, right color=transparent!0]
\tikzfading[name=fade top,
  bottom color=transparent!0, top color=transparent!100]
  
\pgfdeclarelayer{bg}    
\pgfdeclarelayer{fg}    
\pgfsetlayers{bg,main,fg}  

\definecolor{participantAcolor}{RGB}{243,98,33}
\definecolor{participantBcolor}{RGB}{0,102,189}
\definecolor{participantCcolor}{RGB}{66,70,50}
\definecolor{cplColor}{RGB}{0,0,0}
\definecolor{preciceColor}{RGB}{ 161, 177, 25}

\definecolor{ci_pantone300}{RGB}{ 0, 101, 189}

\definecolor{ci_pantone301}{RGB}{ 0, 82, 147}
\definecolor{ci_pantone540}{RGB}{ 0, 51, 89}

\definecolor{ci_pantone283}{RGB}{ 152, 198, 234}
\definecolor{ci_pantone542}{RGB}{ 100, 160, 200}
\definecolor{ci_ivory}{RGB}{ 218, 215, 203}
\definecolor{ci_orange}{RGB}{ 227, 114, 34}
\definecolor{ci_green}{RGB}{ 162, 173, 0}